\documentclass{article}
\usepackage[final]{colm2026_conference}
\usepackage{microtype}
\usepackage{hyperref}
\usepackage{url}
\usepackage{booktabs}
\usepackage{lineno}
\definecolor{darkblue}{rgb}{0, 0, 0.5}
\hypersetup{colorlinks=true, citecolor=darkblue, linkcolor=darkblue, urlcolor=darkblue}

\usepackage[T1]{fontenc}
\usepackage[utf8]{inputenc}
\usepackage{tcolorbox}
\tcbuselibrary{breakable}
\tcbuselibrary{skins}
\usepackage{ifthen}
\usepackage{wrapfig}

\newcommand{\benchnllb}{\textsc{NLLB}}
\newcommand{\benchflores}{\textsc{FLORES}}
\newcommand{\benchglotlid}{\textsc{GlotLID}}
\newcommand{\benchchrfpp}{\textsc{chrF++}}
\newcommand{\benchpolywrite}{\textsc{PolyWrite}}
\newcommand{\benchmhumaneval}{\textsc{mHumanEval}}
\newcommand{\benchmmlu}{\textsc{MMLU}}
\newcommand{\benchmmluredux}{\textsc{MMLU-Redux}}
\newcommand{\benchdylan}{\textsc{DyLAN}}

\newcommand{\modelgpt}{\textsc{GPT-5.4}}
\newcommand{\modelopus}{\textsc{Claude Opus 4.6}}
\newcommand{\modelgemini}{\textsc{Gemini 3.1 Flash-Lite}}
\newcommand{\modelmaverick}{\textsc{Llama 4 Maverick}}
\newcommand{\modelqwen}{\textsc{Qwen 3.5 397B}}
\newcommand{\modelgptmini}{\textsc{GPT-4.1 mini}}
\newcommand{\modelhaiku}{\textsc{Claude Haiku 4.5}}
\newcommand{\modelgeminiflash}{\textsc{Gemini 2.5 Flash}}
\newcommand{\modelllamaoneb}{\textsc{Llama 3.2 1B}}
\newcommand{\modelqwenthreeb}{\textsc{Qwen 2.5 3B}}
\newtcolorbox{appendixexamplebox}[1]{
    breakable,
    colback=teal!6,
    colframe=teal!70!black,
    coltext=black,
    title={#1},
    fonttitle=\bfseries,
    fontupper=\footnotesize,
    coltitle=white,
    colbacktitle=teal!70!black,
    boxrule=0.4mm,
    arc=2mm,
    top=1mm,
    bottom=1mm,
    left=1mm,
    right=1mm,
    boxsep=1.2mm,
    before skip=4pt,
    after skip=4pt
}

\newtcolorbox{takeawaybox}{
  enhanced,
  colback=teal!10,
  colframe=teal!70!black,
  boxrule=0pt,
  leftrule=4pt,
  rightrule=0pt,
  toprule=0pt,
  bottomrule=0pt,
  sharp corners,
  fontupper=\small,
  before skip=6pt,
  after skip=6pt,
  top=4pt,
  bottom=4pt,
}

\title{Lost in the Tower of Babel: The Adverse Effects of Incidental Multilingualism in LLMs}

\author{Anjishnu Mukherjee, Chutong Meng, and Antonios Anastasopoulos \\
  Department of Computer Science \\
  George Mason University \\
  \texttt{\{amukher6,cmeng2,antonis\}@gmu.edu}
  }

\begin{document}

\ifcolmsubmission
\linenumbers
\fi

\maketitle

\begin{abstract}
    This paper argues that contemporary multilingual NLP has converged on a fragile and misleading paradigm of \emph{incidental multilingualism}. Today's LLMs appear multilingual largely because they are trained on massive, uneven web corpora, not because multilingual or multicultural competence has been treated as a core design objective. We contend that this paradigm systematically produces unequal, brittle, and opaque behavior across languages, with severe consequences in real-world and agentic deployments where models must reason, plan, and act across multiple linguistic contexts. We report a focused empirical study of two practical questions: which languages models self-report as supported and which languages they actually respond in across multilingual prompts. We additionally demonstrate how even a simple language-change attack can surface these failures and expose hidden assumptions about language in LLM-based systems. To address this, we call for a shift toward \emph{multilingualism by design}: a research agenda that treats equitable multilingual performance, cultural grounding, and cross-lingual behavioral understanding as first-class goals in all aspects of the model pipeline.\footnote{Results, code and data: \url{https://github.com/antonisa/llm-languages}.}
\end{abstract}

\section{Introduction}

Despite persistent and well-documented gaps, the success of large language models (LLMs) in multilingual settings is undeniable.
Contemporary LLMs exhibit usable competence in over 100 languages and are increasingly deployed in real-world applications across the globe.
This apparent breadth has led to a widespread perception that multilingual language modeling is a largely solved or at least steadily improving problem.

But the same amazing LLMs exhibit embarrassingly problematic limitations. One notable such limitation is their inability to consistently generate text in a user's desired language.
Even if top-tier models such as \textsc{GPT-4o} score between $96$ and $98\%$ on the Language Confusion Benchmark~\citep{marchisio-etal-2024-understanding}, a system that receives more than $2$ billion queries per day (like \textsc{ChatGPT} does), effectively still responds in the wrong language $40+$ million times per day.

In this paper, we argue that this perception of success is somewhat  misleading.
Much of the recent progress in multilingual LLMs rests on shaky foundations that risk constraining future advances and obscuring systemic weaknesses.
In particular, we contend that the dominant paradigm in multilingual LLM development relies on a set of implicit assumptions that deserve renewed scrutiny.
Without revisiting these assumptions, further progress toward robust, equitable, and predictable multilingual systems may prove difficult.

We call multilingual capabilities of current LLMs that emerge largely as a side effect of other decisions, such as training on massive, heterogeneous web corpora, \textit{incidental multilingualism}. We contrast it with \textit{multilingualism by design}, where multilingual capability is treated as a primary design goal, and explicit objectives and success criteria across languages and cultures shape data, modeling, evaluation, and deployment. 
Our experiments aim to diagnose behavioral symptoms and support gaps that might contradict even the training intent of LLM providers. To ground this argument in evidence relevant for deployment, we analyze a two-way support gap between self-declared support lists produced by models when asked directly, and observed non-refusal multilingual behavior under diverse task prompts.

This paper makes four contributions. First, we briefly situate our argument within the current status quo of multilingual language modeling (\S\ref{sec:statusquo}). Second, we present a focused investigation of language support claims versus observed behavior across current frontier models (\S\ref{sec:what_languages}). Third, we demonstrate how a simple language-change ``attack'' exposes failure modes of incidental multilingualism in practical LLM-powered agentic scenarios (\S\ref{sec:tob}). Finally, we outline a set of research directions and policy suggestions aimed at moving the field toward multilingualism by design (\S\ref{sec:discussion}).

We aim for this paper to be neither an exhaustive survey nor an explicitly empirical one.
Rather, it is a position piece grounded in concrete investigation and examples, intended to provoke discussion and re-examination of prevailing practices in multilingual LLM research.

\section{The Status Quo in Multilingual NLP}
\label{sec:statusquo}

Early massively multilingual modeling was driven by research in machine translation~\cite[e.g.,][]{aharoni-etal-2019-massively} and eventually hit the spotlight with encoder-only transformer language models such as multilingual BERT (mBERT) and XLM-R. mBERT~\citep{devlin-etal-2019-bert} was trained on Wikipedia in 100+ languages with a shared subword vocabulary, the first model that enabled zero-shot transfer across languages but also exhibiting strong performance disparities between high- and low-resource languages \citep{wu-dredze-2020-languages}. Generative LLMs such as GPT-$n$ and various open-source families like LLaMA~\citep{touvron2023llama2} inherit this \emph{incidental} multilingualism: they are trained on massive but heavily skewed web corpora, which gives them surprisingly broad language coverage but without explicit design for facilitating language transfer or equitable performance across languages~\citep{doddapaneni2021primer,qin2025survey}.
Below, we outline general observations that we believe hold for the vast majority of multilingual LLM research and development.

\noindent \textbf{Uneven Performance and Evaluation Issues}
A consistent finding across models and tasks is that multilingual performance is highly imbalanced and, in many cases, ``unfair''.
Accuracy can vary dramatically across languages even when tokenization and architecture are shared, with languages that are low-resource or typologically distant from English lagging behind.
This was true for mBERT~\citep{pires-etal-2019-multilingual} and still is to this day for the most recent models~\citep{song-etal-2023-globalbench}. Even worse, \citet{ahuja-etal-2024-megaverse} found robust indicators that several models are likely to be contaminated with multilingual evaluation benchmarks, making proper evaluation at scale almost impossible.

Much recent multilingual work focuses on evaluation and dataset construction to quantify these problems.
Benchmarks such as XTREME~\citep{ruder-etal-2021-xtreme}, XGLUE~\citep{liang-etal-2020-xglue}, \benchflores{}~\citep{goyal-etal-2022-flores}, MIRACL~\citep{zhang-etal-2023-miracl}, and multilingual safety benchmarks like XSafety~\citep{wang-etal-2024-languages} provide multi-task, many-language test suites that make cross-lingual gaps and safety issues more visible.
Efforts like GlobalBench~\citep{song-etal-2023-globalbench}, CultureBench~\citep{chiu-etal-2025-culturalbench}, or DialectBench~\citep{faisal-etal-2024-dialectbench} aimed to organize benchmarking around specific goals (holistic LM evaluation, cultural appropriateness, and robustness to dialectal variation respectively).

The Aya initiative did push toward ``multilingualism by design,'' at least on the data side: the Aya Dataset and Collection~\citep{singh-etal-2024-aya} constitute large-scale multilingual instruction corpora with deliberate coverage of 100+ languages, and Aya-23~\citep{aryabumi2024aya23} traded breadth for depth to improve quality in 23 languages, releasing open-weight models and detailed multilingual evaluation to track progress~\citep{cohere2024languagegap}.

In parallel, data- and tokenizer-centric efforts aim to mitigate structural disadvantages.
The argument can be made that byte- or character-level models such as ByT5~\citep{xue-etal-2022-byt5} remove language-specific subword vocabularies and show competitive performance across scripts, with follow-up work demonstrating that tokenization decisions materially affect low-resource and morphologically rich languages \citep{edman2024character,dang2025morphology}.
More recent efforts even attempt to go tokenization-free by converting text inputs to images, in the PIXEL model series~\citep{kesen-etal-2025-multilingual, tatariya-etal-2024-pixology}.

Despite this intense activity around benchmarks, data curation, and tokenization, there is comparatively less work on \emph{understanding} what massively multilingual (200+ language) models actually know and how languages interact internally.
Unfortunately, properly understanding such complicated systems is highly computationally expensive: \citet{malkin-etal-2022-balanced} rely on approximations from bilingual modeling, while \citet{faisal-anastasopoulos-2024-efficient} similarly relied on bi- and trilingual adapters on top of pre-trained multilingual models.
Surveys emphasize that most analyses still focus on surface-level performance or a small subset of languages, leaving open questions about latent language clustering, cross-lingual interference, knowledge sharing, and how cultural and moral biases propagate across languages in very large models \citep{doddapaneni2021primer,qin2025survey,hammerl2022moral,jain2024culture}.

A telling sign of the field’s current state, and of why our paper is still very much needed, is that we are still restating critiques made more than a decade ago. \citet{bender-2009-linguistically} warned against treating systems as language-independent simply because they are linguistically naïve. \citet{bender2011achieving} further argued that the field was neither systematically evaluating language-independence nor building it with sufficient linguistic sophistication. Later work revisited the same structural problem from different angles: popular cross-lingual proxy evaluations can misrepresent downstream generalization \citep{glavas-etal-2019-properly}, the English-centric allocation of research effort reproduces inequality \citep{sogaard-2022-ban}, and much of NLP still defaults to a narrow ``square one'' paradigm centered on English and accuracy alone \citep{ruder-etal-2022-square}. That these critiques remain salient in the LLM era suggests that multilinguality has too often been treated as an emergent byproduct of scale rather than as an explicit design objective.

In this sense, the field is in a transitional phase: we can increasingly measure and partially patch inequities in multilingual LLMs. And this is uncontroversially a crucial, necessary first step. But fully explaining how 100-200+ languages co-exist and shape one another inside these models in terms of both representations and socio-cultural behavior remains an important open research frontier.

\noindent \textbf{Consequences in Real-World Settings}
The reliance on \emph{incidental multilingualism} has direct and often problematic consequences for real-world deployment of LLMs. In principle, multilingual systems deployed in domains such as education, healthcare, public services, or customer support should offer comparable reliability, safety, and calibration across languages. In practice, however, performance and safety lack data availability and research attention. Consequently, what is often presented as a single global model behaves like a collection of language-conditioned systems with divergent capabilities and failure~modes.

These disparities are particularly pronounced in safety and cultural adequacy. LLMs generate significantly more unsafe responses for non-English prompts~\citep{wang-etal-2024-languages}, and multilingual jailbreaks exploiting low-resource languages are easier to construct and harder to defend against \citep{shen2024languagebarrier}. At the same time, multilingual models are not inherently multicultural~\citep{adilazuarda-etal-2024-towards}. They frequently encode Anglocentric norms even when operating in other languages, leading to culturally misaligned or inappropriate behavior. LLM safety evaluation and mitigation efforts overwhelmingly focus on English, leaving safety guarantees in other languages underdeveloped \citep{yong2025state}.
In deployment, this translates into higher risks precisely for the most vulnerable users with the least institutional oversight or recourse.

The growing use of LLM-powered \emph{agentic systems} amplifies these concerns.
Such systems must interpret instructions, retrieve information, and act across multiple linguistic contexts, yet even monolingual agent reliability remains an open research problem \citep{cheng2024exploring}.
Introducing multilinguality creates additional failure modes~\citep{chen-etal-2024-impact}. Each language boundary potentially enables semantic drift, misalignment, or unsafe actions, underscoring the risks of deploying agentic systems built on incidental multilinguality.

\noindent \textbf{From Ideal Design to Current Practice}
Taken together, the previous notes highlight a growing gap between an ``ideal'' multilingual-by-design ecosystem and current practice.
In the ideal picture, models and agents would be trained and evaluated with explicit language- and culture-aware objectives; safety mitigations, datasets, and benchmarks would be co-developed across languages; and real-world systems would guarantee comparable reliability and protections regardless of users' background. By contrast, most deployed LLMs today still rely on English-centered safety tooling. This tension is particularly acute for agentic systems, where multilingual failures can propagate into actions rather than remaining as text. Bridging this gap will require a shift toward treating multilingualism and multicultural competence as central design goals.

\section{What Languages Do LLMs Speak?}
\label{sec:what_languages}

In practice, ``language support'' is not a single property. We offer here at least three notions, which can diverge sharply:
$(1)$ the languages a model claims to support when asked directly,
$(2)$ the languages in which it can produce text that is actually identified as the target language, and
$(3)$ the languages in which it can perform nontrivial downstream tasks reliably.\footnote{Of course, there's also the notion of ``official'' support by way of listing languages in model cards or release notes. But no major provider provides such a list.}
When these notions disagree, both users and deployers lose a stable operational definition of support.
This distinction matters in high-stakes settings such as education, public services, healthcare, and legal assistance, where deployment decisions are often justified by broad multilingual claims rather than by task-specific evidence.
Our question in this section is therefore deliberately operational: when we move from self-description to behavior, what languages do current frontier LLMs actually ``speak''?

\subsection{Experimental Design}

We evaluate five frontier LLMs: \modelgpt{}, \modelopus{}, \modelgemini{}, \modelmaverick{}, and \modelqwen{}. Our goal is to measure both what languages a model claims to support (\emph{declared support}) and how it behaves on downstream tasks in those languages (\emph{behavior}).

\paragraph{Declared support}
We first study how models describe their own multilingual coverage.

$(1)$ \emph{Elicitation.} Each model receives $20$ English prompt variants asking for a JSON list of supported languages and ISO codes (Figure~\ref{fig:section3_support_probe}).

$(2)$ \emph{Verification.} Since provider documentation for these models does not give a complete language-by-language ground-truth list, we cannot score these support lists directly. Instead, for each of the $20$ elicitation runs, we ask the model to generate a sentence in every language it claimed in that specific run. Thus, verification is repeated once per claimed language per prompt variant. We then pass each generated sentence through \benchglotlid{} \citep{kargaran-etal-2023-glotlid} to confirm whether the detected language matches the claimed one. Figure~\ref{fig:section3_support_probe} reports intersection ($I$) and union ($U$) bounds over both the raw language sets and the confirmed language sets across the $20$ prompts.

$(3)$ \emph{Language support awareness.} We next evaluate two additional tasks on each model's LangID-confirmed supported-language union bound $U$. First, we provide the model with a sentence in one of these confirmed languages and ask it to identify the language code (\emph{lang\_id}). Second, we provide the name of that language and ask the model to answer \emph{yes} or \emph{no} when asked whether it can write or speak it (\emph{language\_self\_report}). If models were well calibrated on their own confirmed support frontier, they should perform near-perfectly on both tasks. We score \emph{lang\_id} by exact correctness of the predicted language code, and \emph{language\_self\_report} by whether the model answers \emph{yes} for languages in its own LangID-confirmed support frontier $U$.

\paragraph{Behavior}
We next evaluate multilingual behavior on downstream tasks (Figure~\ref{fig:section3_takeaways}). These task families use a shared $203$-language collection: English along with the $202$ non-English languages in a \benchflores{}/\benchnllb{}-backed translatable set. Non-English prompt translations are generated with \benchnllb{}-$200$ $3.3$B \citep{nllbteam2022languageleftbehindscaling}. Appendix~\ref{sec:measurement_audits} audits NLLB prompt quality with COMETKiwi~\citep{rei-etal-2022-cometkiwi} and GlotLID on known-language reference samples.

$(1)$ \emph{Short story generation.} We ask the model to generate a culturally relevant short story of roughly $4$-$6$ sentences in the target language.

$(2)$ \emph{Teacher conversation.} We assign the user the persona of a language learner and ask the model to hold a short pedagogical conversation in the target language.

For both of these tasks, we evaluate whether the output is in the requested language using \benchglotlid{}.

$(3)$ \emph{Translation to/from English.} We ask models either to translate text into English or from English into the target language while preserving meaning and tone. We evaluate these tasks in two ways: target-language retention is measured with \benchglotlid{}, and translation quality is measured with \benchchrfpp{} \citep{popovic-2017-chrf}.

$(4)$ \emph{Code generation.} We describe a programming task in natural language and ask the model to generate Python code. We evaluate this task against \benchmhumaneval{} \citep{raihan-etal-2025-mhumaneval}. We use \emph{parseability}, indicating whether the generated code can be parsed into an abstract syntax tree, and \emph{functionality}, indicating whether the generated code passes the benchmark unit tests, as our metrics.

$(5)$ \emph{Long-form generation.} Last, we evaluate open-ended generation using \benchpolywrite{} \citep{ji2025emma500enhancingmassivelymultilingual}. Although this benchmark is available for $240$ languages, only $141$ overlap with the \benchflores{} collection used in our other tasks (and we only report results on this subset).

\subsection{Results and Discussion}

For our results below, we report performance grouped by the language resource classes of \citet{joshi2020statefate}, which classified languages according to their resource availability. The downstream task families are aggregated over repeated runs of the current evaluation pipeline, and we summarize model performance with the task-specific metrics above. We also explicitly permit models to refuse when they cannot complete a task reliably. In practice, however, the dominant failure mode is usually incorrect generation rather than refusal.

\begin{figure*}[t]
    \centering
    \includegraphics[width=.8\textwidth]{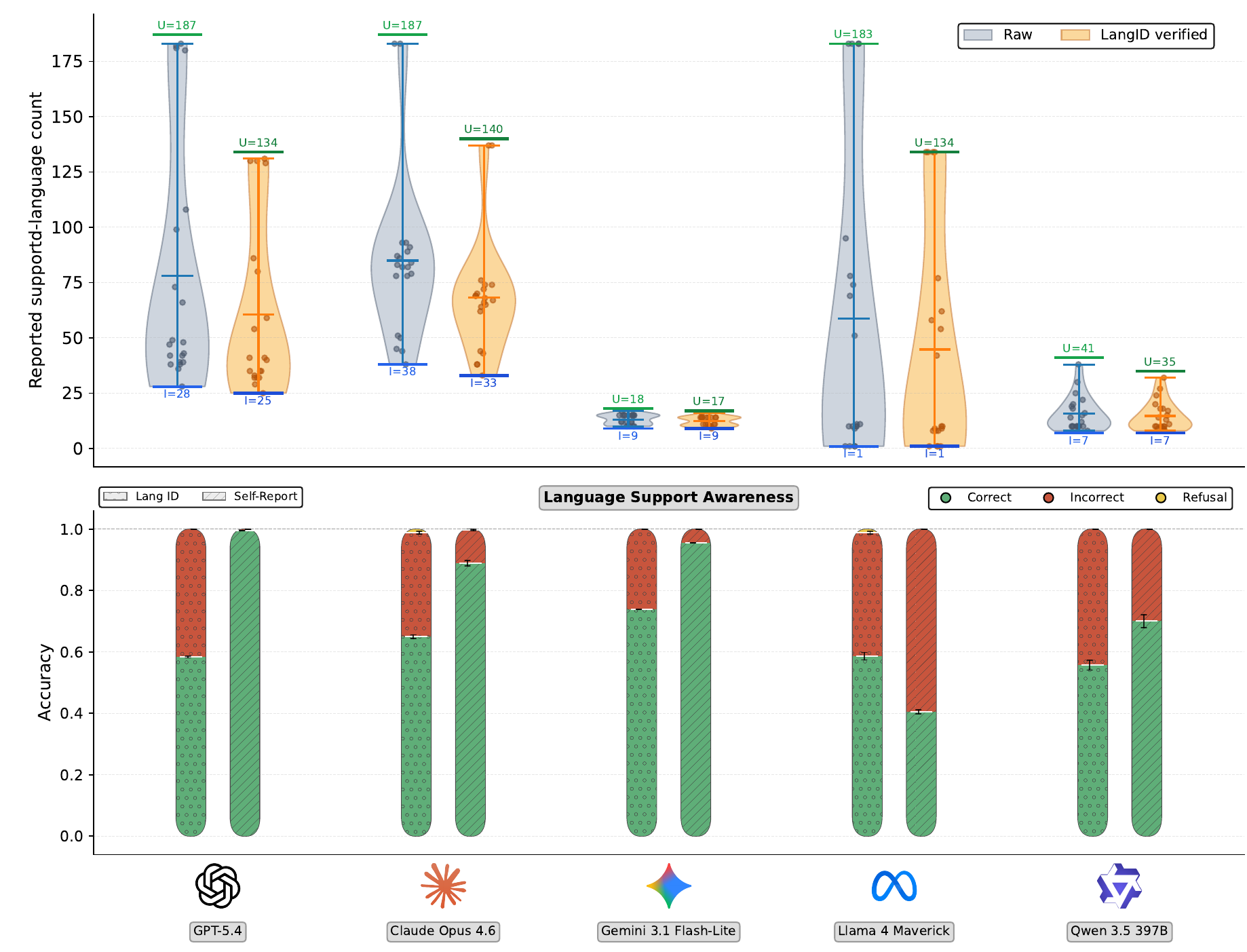}
    \vspace{-1em}
    \caption{Language support claims are highly prompt sensitive (top, \emph{declared support}) and are not well calibrated on each model's confirmed support frontier (bottom, \emph{behavior}). The upper panel shows how much the set of supposedly supported languages changes across $20$ English paraphrases of the same question. The lower panel evaluates language identification and direct yes/no self-report only on languages that are LangID verified for that model.}
    \label{fig:section3_support_probe}
    \vspace{-1em}
\end{figure*}

\paragraph{Declared support}
Figure~\ref{fig:section3_support_probe} and Table~\ref{tab:section3_support_probe} show that the multilingual support frontier exposed by these models is neither stable nor well calibrated.

$(1)$ \emph{Elicitation.} Prompting the same support-list question in slightly different paraphrases, yields a wide spread of claimed support frontiers. Raw support unions range from just $18$ languages for \modelgemini{} to $187$ for both \modelgpt{} and \modelopus{}, with \modelmaverick{} reaching $183$. Even within a single model, counts vary sharply across paraphrases, for example \modelmaverick{} ranges from $1$ to $183$.

$(2)$ \emph{Verification.} After verifying the sentences generated for each language the models claim to support, the union $U$ decreases for every model, more for some and less for others. Raw self-declared support is thus consistently too optimistic, and verification makes the behavioral frontier substantially more conservative. Low intersection $I$ is an even bigger problem. Even after verification, confirmed intersections remain small relative to the union upper bound. This means many languages are only intermittently acknowledged as supported across paraphrases. \modelmaverick{} suffers the most from this phenomenon, as its confirmed union is broad ($U=134$), yet only one language (English) survives across all $20$ prompt variants and reruns.

\begin{takeawaybox}
\textbf{Takeaway $\mathbf{1}$:} Declared multilingual support is unstable under simple prompt variation, and LangID verification reduces the set of languages models can plausibly claim to support.
\end{takeawaybox}

$(3)$ \emph{Language support awareness.} Table~\ref{tab:section3_support_probe} shows that the \emph{language\_self\_report} task is generally easier for models than target language identification (\emph{lang\_id}), implying that these models claim support in languages where they cannot reliably identify text. These results support our call for much stronger provider transparency: models should expose versioned language-support frontiers and explicitly decline languages that fall outside them.

\begin{figure*}[t]
    \centering
    \includegraphics[width=.75\textwidth]{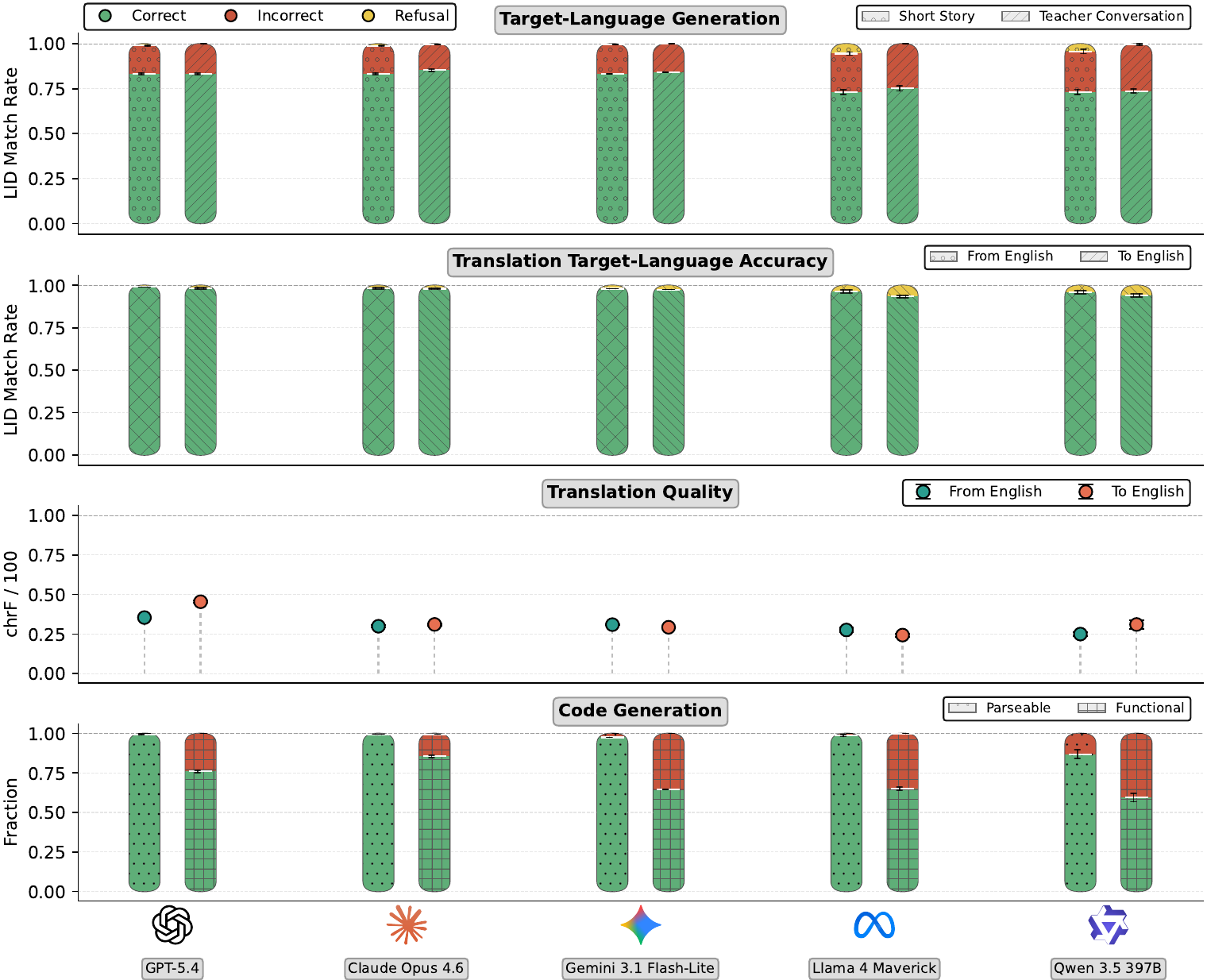}
    \vspace{-1em}
    \caption{Multilingual task performance on $203$-languages across $4$ tasks - Short stories and pedagogical dialogues test whether models can generate in the requested language, the translation panels separate ability to generate in target-language from translation quality measured with \benchchrfpp{}, and the code generation panel illustrates the difference between parseable Python and functionally correct code on multilingual \benchmhumaneval{} prompts.}
    \label{fig:section3_takeaways}
    \vspace{-1em}
\end{figure*}

\begin{figure*}[t]
    \centering
    \includegraphics[width=.65\textwidth]{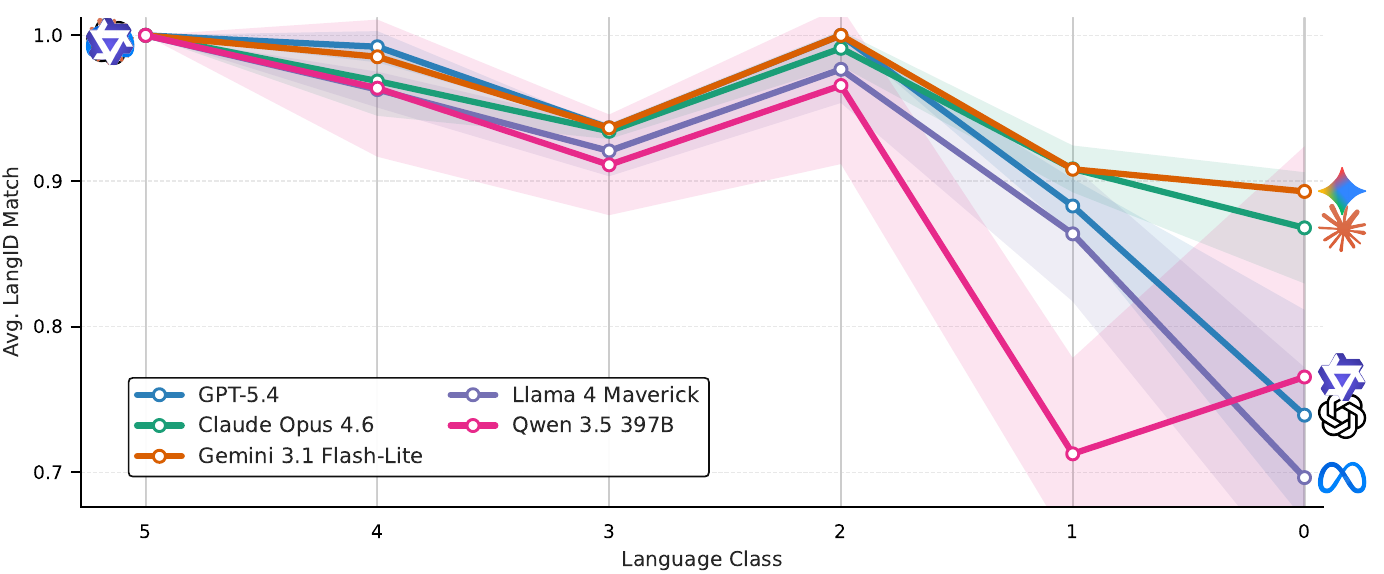}
    \vspace{-0.5em}
    \caption{Target-language retention in long-form writing across $141$ languages, grouped by six resource classes~\citep{joshi2020statefate}. Class~$0$ contains languages with the least language resources and class~$5$ the most. The hardest classes widen the gaps between models.}
    \label{fig:section3_taxonomy}
    \vspace{-1em}
\end{figure*}

\paragraph{Behavior}
Figures~\ref{fig:section3_takeaways} and \ref{fig:section3_taxonomy}, along with Table~\ref{tab:section3_multilingual_results}, show that multilingual behavior is strongly task-dependent, and that incorrect generation is far more common than explicit refusal.

$(1)$ \emph{Short story generation} and $(2)$ \emph{teacher conversation.} Short-form target-language generation is comparatively close across models. Most are generated in the requested language, but the open-source models are the only ones that regularly refuse requests.

$(3)$ \emph{Translation to/from English.} \modelgpt{} leads both translation \benchchrfpp{} directions ($35.44$ EN$\rightarrow$X and $45.46$ X$\rightarrow$EN). Translation quality and generating in the requested language measure different abilities, and the results show that generating in the correct language does not always imply a high-quality translation.

$(4)$ \emph{Code generation.} \modelopus{} leads code functional correctness ($0.855$), while parseable code is generally easier to generate for all models than functionally correct code.
This gap shows that models can often exploit strong Python priors to produce syntactically valid code even without fully capturing the task requirements.

For these four downstream tasks, non-refusal rates typically remain above $0.94$ and often above $0.98$, while correctness is much lower. So where models should arguably abstain, they instead produce incorrect language output, poor translations, or non-functional code. This matters operationally because the potential user sees an apparently cooperative model rather than a transparent admission of weak support.

\begin{takeawaybox}
\textbf{Takeaway $\mathbf{2}$:} Across downstream multilingual tasks, the dominant failure mode is usually incorrect generation rather than explicit refusal.
\end{takeawaybox}

$(5)$ \emph{Long-form generation.}
Figure~\ref{fig:section3_taxonomy} shows clear differences across models as we progress from high-resource languages to lower resourced buckets. All models saturate class~$5$ and remain close in classes~$2$-$4$, but have distinctly separate performance for classes~$0$-$1$. So the most meaningful differences between models appear not in high-resource languages, where all do well, but in low-resource settings where replying in the requested language is much less reliable and broad claims about coverage fail to predict actual usability.

\begin{takeawaybox}
\textbf{Takeaway $\mathbf{3}$:} The hardest low-resource languages remain the clearest separator between models, and broad support claims are least predictive exactly where multilingual robustness matters most.
\end{takeawaybox}

\paragraph{Discussion}

These empirical results argue against treating multilingual support as a binary model attribute.
At minimum, reporting should separate \emph{declared support}, \emph{stable support}, and \emph{task-conditional behavioral support}.
Otherwise a claim such as ``supports 100+ languages'' collapses over phenomena that are empirically distinct and operationally consequential.

Our comparison of frontier models also suggests that there is no single multilingual ordering.
\modelgpt{} leads in translation, \modelopus{} in multilingual code correctness, \modelgemini{} in target-language generation for the hardest \benchpolywrite{} buckets, while \modelmaverick{} and \modelqwen{} are competitive on some mid-resource slices despite weaker support-list stability.
Multilingualism is therefore not a monolithic capability that can be summarized by one headline number.
It is a structured capability profile, with different models tracing out different frontiers across support claims, translation, language retention, and long-tail generation.

\section{The Tower of Babel Problem}
\label{sec:tob}

Recent work increasingly explores \emph{collaborative} and \emph{agentic} LLM systems, in which multiple language-model-based agents interact, and jointly produce outputs.
A representative example is \benchdylan{}~\citep{liu2024dynamic}, where several agents (instantiated from the same underlying LLM) independently propose solutions, evaluate one another's responses, and are dynamically selected to collaborate through iterative discussion rounds.
Such systems implicitly assume that communication between agents is semantically stable and that linguistic variation does not fundamentally
alter reasoning, coordination, or task performance.

This assumption though is fragile in the presence of incidental multilingualism.
Modern LLMs exhibit language-dependent behavior, performance, and alignment properties.
In some cases, such variation may be appropriate, for example, when agents are expected to adapt to linguistic or cultural context. However, agentic systems that operate across languages are vulnerable to failures that arise purely from linguistic heterogeneity rather than task complexity.
We refer to such failures as the \textbf{Tower of Babel (ToB) problem}: a scenario in which otherwise identical agents are induced intentionally or accidentally to communicate in different languages, thereby degrading coordination and task success.

It would be straightforward for an adversary to devise a ToB attack, exploiting the fact that LLM behavior is not invariant across languages.
The intuition is that in practical deployments, agents may interact with heterogeneous external data sources, user inputs, or retrieved documents in multiple languages.
This creates a natural surface for accidental or adversarial language shifts through multilingual web content, translation artifacts, or data poisoning or prompt injections that nudges agents to respond in specific languages.
Under incidental multilingualism, there is no guarantee that agents operating in different languages will reason, evaluate, or communicate in mutually compatible ways, even when they share the same underlying model.

\begin{wrapfigure}[6]{r}{.5\textwidth}
    \small
    \centering
    \vspace{-2em}
    \begin{tabular}{@{}c|cc@{}}
    \toprule
         & all English & ToB attack \\
    \midrule
        \benchmmlu{} accuracy & $75.0$ & $67.5$ \\
    \bottomrule
    \end{tabular}
    \vspace{-.5em}
    \caption{Task performance degrades substantially when agents operate multilingually.}
    \label{tab:mmlu}
\end{wrapfigure}

\paragraph{Illustrative proof of concept}
As a demonstration, we simulate the ToB issue in a simplified collaborative-agent setting.
Similar to \citet{liu2024dynamic}, we instantiate seven role-based agents (\texttt{Economist}, \texttt{Doctor}, \texttt{Lawyer}, \texttt{Mathematician}, \texttt{Psychologist}, \texttt{Programmer}, \texttt{Historian}) and evaluate them on \benchmmluredux{}~\citep{gema-etal-2025-done}.
We compare two conditions: (a) a monolingual baseline where all agents communicate in English, and (b) a multilingual condition where each agent is tricked into operating in a different language (English, Mandarin Chinese, isiZulu, Greek, Hungarian, Bengali, Quechua, Turkish, or Romanian).\footnote{Languages are chosen mostly for geographic and typological diversity.}
Aside from language, the setup is identical across conditions.

\paragraph{Second-order effects}
The \benchdylan{} framework relies on \emph{inter-agent evaluation}: each agent scores the initial responses of the others, and these scores determine which agents are selected to participate in the collaborative refinement stage.
In principle, these evaluations should reflect the substance, correctness, and usefulness of each response, independent of the language in which it is expressed.

But, we observe language-dependent differences in these scores.
For example, Greek-speaking agents consistently assign Turkish-speaking agents scores approximately one standard deviation lower than those assigned to agents operating in Western European languages (and vice versa).
Similar patterns emerge (to a smaller extent) between the Romanian and Hungarian agents.\footnote{To be fair, our choice of several Balkan languages might lead to exaggerated effects due to the region's history. But it certainly supports our point about unintended effects.}
Perhaps unsurprisingly, agents operating in Quechua and isiZulu receive substantially lower scores from nearly all other agents, regardless of task content.
While these descriptive patterns are anecdotal and some of this effect may be attributable to limited or uneven model fluency in these languages, the key observation is that language alone exerts a strong influence on agent selection dynamics.

\paragraph{Implications}
The performance drop in Figure~\ref{tab:mmlu} occurs without model or task changes and is consistent with language-induced variation in agent behavior, although our design does not explicitly separate individual comprehension from coordination effects.
While simple, the result illustrates how incidental multilingualism may undermine \emph{performance} and \emph{coordination} in agentic LLM systems.
Language-dependent biases in inter-agent evaluation suggest a possible second-order failure mode: agents operating in certain languages are systematically devalued regardless of the quality of their contributions.
In real-world deployments, this risks reproducing linguistic and cultural hierarchies inside ostensibly neutral AI systems, where language choice implicitly shapes authority, trust, and influence.

\section{Toward Multilingualism by Design: Some Concrete Directions}
\label{sec:discussion}

The reader is hopefully convinced by now that research must treat multilingual capability as an explicit design objective across the full LLM pipeline.
Below we offer a set of concrete directions that go beyond the usual calls for curated data selection, balanced data mixtures, or calls for extensive multilingual evaluation and culturally grounded benchmarks.\footnote{Not because these are not needed -\textit{au contraire!}- but because they do not challenge the fundamental issues this paper aims to highlight.}

At the core of multilingual language modeling lies a fundamental tension between \emph{language-agnostic representation learning} and \emph{language- and culture-specific behavior}.
Language agnosticism enables cross-lingual transfer and parameter sharing.
But language- and culture-specificity is essential for appropriate reasoning, generation, and interaction in real-world contexts.
Strongly shared representations such as those induced by joint subword vocabularies and fully shared transformer parameters are the current paradigm, but they also risk collapsing meaningful linguistic, cultural, and pragmatic distinctions.
In many settings, it is absolutely fine if ``LLMs reason in English''~\citep{etxaniz-etal-2024-multilingual}, but in many others it would be catastrophic.
In our view, the translate-test and in general translation-based approaches~\citep{liu-etal-2025-translation,ebrahimi-wense-2024-zero} are problematic fixes to a persistent problem that can never become an appropriate general solution regardless of how good translation capabilities become.

Conversely, introducing language-specific components or objectives can preserve local variation, but may reduce transfer efficiency and complicate scaling, a trade-off that remains poorly understood.
Somewhat-forgotten work from the encoder LLMs era on explicit alignment objectives~\citep{hu-etal-2021-explicit} suggests that this trade-off is not binary, especially if put into an envisioned context of more carefully curated multilingual data at scale such as the Aya efforts~\citep{singh-etal-2024-aya}.
Careful design choices should improve cross-lingual alignment without erasing language-specific signals, and even drive overall performance up. Or at a minimum, the system design should be based on a pre-specified target along this tension continuum.

However, practitioners still lack principled guidance on how and where this balance should be managed. Different interventions at the level of tokenization, intermediate representations, training objectives, decoding, or post-tuning have been explored. These contributions are certainly steps in the correct direction, but substantial progress requires that they are perhaps employed in combination instead of as possible one-off, often post-hoc fixes.

Last, we argue that more work is needed in properly understanding multilinguality as reflected in LLMs. Our models appear to have both language-invariant components~\citep{foroutan-etal-2022-discovering} and language-specific subnetworks~\citep{tang-etal-2024-language}, and both of these are crucial for different aspects of performance.
This is perhaps unsurprising, but the mechanisms through which such components arise internally, how they might be shaped by cross-language interactions (be that positive transfer or negative interference), or how to identify them so as to study them, are all entirely open questions.

In the more specific context of collaborative LLM-powered \textit{agents}, we note that current setups are fundamentally constrained by the requirement to communicate and coordinate exclusively through human language.
Human languages are a medium shaped by human communicative needs and embodied assumptions rather than by the requirements of machine reasoning or coordination.
While human languages provide an accessible interface for interaction, they may also introduce ambiguity or cultural biases
when used as the sole channel for inter-agent communication.
This raises the question of whether
agentic systems should rely entirely on natural language internally, or whether alternative, non-linguistic or hybrid communication representations designed for stability, precision, or cross-lingual invariance could better support robust coordination in multilingual settings.
Early work explored machine-to-machine communication~\citep{tebbifakhr-etal-2019-machine}, as has the field of emergent communication~\citep{cao2018emergentcommunicationnegotiation}, but such approaches have not found fertile ground in the current practical LLM-dominated environment.
Exploring such representations might be a promising direction for decoupling agent reasoning and coordination from the incidental, often unwanted baggage that human languages carry.

\paragraph{Provider transparency and accountability}
We also advocate for explicit multilingual accountability standards at release time. Providers should publish versioned supported-language lists, together with language-level evaluation cards that report both quality and safety metrics rather than broad headline claims of multilingual support.
They should further document language-conditioned refusal policies, making clear when support is considered inadequate and when the model is expected to abstain instead of improvising.

When a model cannot reliably support a language, it should transparently refuse rather than produce overconfident low-quality outputs.
Such refusals should be paired with safer fallback guidance: for example, suggesting a better-supported language, requesting clarification, or directing the user toward a human escalation path where appropriate.
This is not a complete substitute for multilingual capability, but it is a concrete harm-reduction mechanism.
In the current paradigm, users are often exposed to hidden failure behind fluent-looking text; refusal-by-design makes those capability boundaries explicit and aligns model behavior more closely with what the system can actually do well.

It is worth pointing out that this is a question requiring hard language-specific decisions. 
The distinctions between languages are never clear cut, and this translates to different results for different tasks. 
For example, a multilingual LLM that is extremely competent in Spanish and Portuguese might be adequately capable of \textit{understanding} and executing an instruction given in Mirandese or Asturian, but woefully inaccurate in actually \textit{generating} fluent Mirandese text output.
Thus, careful consideration should be given into whether a language like Mirandese should be included in the language support or whether any refusals should be task-specific.
We do have a concrete answer to how one should balance the potential harms from the potential utility a language speaker might receive, but LLM providers should at least be transparent about the decisions that guide their products.

\section{Conclusion}

Modern LLMs exhibit broad multilingual behavior, but our findings show that official support claims, self-declared support lists, and observed multilingual behavior can diverge sharply.
This mismatch is a direct symptom of incidental multilingualism: language capability emerges, but is not specified as a robust contract.
Moving toward multilingualism by design requires explicit support definitions, multilingual evaluation transparency, and safe refusal behavior when support is inadequate.

\section*{Acknowledgments}
We are grateful to the anonymous reviewers and area chairs from both COLM and ARR whose feedback improved the paper.
Part of this work (the question of which languages frontier LLMs actually support) was the result of thoughtful discussions with Yulia Tsvetkov, Sunayana Sitaram, and Monojit Choudhury.
This work is financially supported by the National Science Foundation under CAREER award 2439202.
\bibliography{paper_references}
\bibliographystyle{colm2026_conference}

\clearpage
\appendix

\section{Task and Prompt Inventory}

This section briefly summarizes the materials behind the multilingual behavior probe.
It is intended to make the evaluation scope concrete for readers who want to see what kinds of prompts and task families produced the reported results.
The support-list probe uses $20$ English prompt variants that differ in wording and scenario framing but always ask for a structured list of supported languages and ISO codes.
The behavioral probe then tests the same models on language identification, direct self-report, short-form generation, translation in both directions, pedagogical dialogue generation, and multilingual code generation.

\begin{appendixexamplebox}{Behavioral task families used throughout Section~3}
\textbf{Language identification.} The model reads a short monolingual snippet and predicts which language it is.

\textbf{Direct self-report.} The model is asked whether it can write or speak a named language.

\textbf{Target-language generation.} The model is asked to produce either a short story or a teacher-student dialogue in the requested language.

\textbf{Translation.} The model translates from English into the target language and from the target language back into English.

\textbf{Code generation.} The model solves multilingual Python programming prompts aligned to \benchmhumaneval{}-style tasks.
\end{appendixexamplebox}

\begin{appendixexamplebox}{Support-list elicitation used for the support frontier analysis}
\textbf{Prompt family.} We use $20$ English prompts spanning direct requests, paraphrases, and light scenario framing.

\textbf{Goal.} Each prompt asks the model to enumerate the languages it supports and to provide language identifiers in a structured format.

\textbf{Why multiple prompts matter.} The support frontier is not treated as a single response but as a stability problem: we compare the intersection and union induced by these $20$ variants, then check which listed languages survive LangID confirmation in downstream behavior.
\end{appendixexamplebox}

\begin{appendixexamplebox}{Support-list elicitation prompts}
\textbf{Representative direct prompt.}
{\ttfamily
List all supported languages with their ISO codes in JSON format.\par
}

\textbf{Representative paraphrase.}
{\ttfamily
What languages do you support? Return a JSON object with language names and ISO codes.\par
}

\textbf{Representative scenario prompt.}
{\ttfamily
For procurement, provide a JSON list of all supported languages plus their ISO codes.\par
}

\textbf{Prompt family note.} The complete elicitation family contains $20$ such English variants, mixing direct requests and scenario framing while preserving the same underlying request.
\end{appendixexamplebox}

\begin{appendixexamplebox}{Prompts for language awareness tasks}
\textbf{Language identification (\emph{lang\_id}).}
{\ttfamily
You are given a text snippet.\par
Return the ISO 639-1 language code only.\par
Text: \{text\}\par
}

\textbf{Direct self-report (\emph{language\_self\_report}).}
{\ttfamily
Can you write or speak \{target\_language\_name\}? Reply only with yes or no.\par
}
\end{appendixexamplebox}

\begin{appendixexamplebox}{Prompts for short-form generation tasks}
\textbf{Short story generation.}
{\ttfamily
Write a short story in \{target\_language\_name\}.\par
Keep it between 4 and 6 sentences.\par
Use simple, natural language and make sure it is culturally-relevant to the \{target\_language\_name\} community.\par
Respond with "unable to perform this task" if you cannot do that.\par
}

\textbf{Teacher conversation.}
{\ttfamily
Can we have a conversation in \{target\_language\_name\}? I am learning \{target\_language\_name\} and I want to practice.\par
}
\end{appendixexamplebox}

\begin{appendixexamplebox}{Prompts for translation and code tasks}
\textbf{Translation to English.}
{\ttfamily
Translate the following text to English.\par
Preserve meaning and tone.\par
Respond with "unable to perform this task" if that's the case.\par
Text: \{text\}\par
}

\textbf{Translation from English.}
{\ttfamily
Translate the following text from English to \{target\_language\_name\}.\par
Preserve meaning and tone. Respond with "unable to perform this task" if that's the case.\par
Text: \{text\}\par
}

\textbf{Code generation.}
{\ttfamily
Generate Python code for the following programming task.\par
Return code only.\par
Task: \{text\}\par
}
\end{appendixexamplebox}

\begin{appendixexamplebox}{Prompt for long-form generation}
\textbf{Long-form generation (\benchpolywrite{}).} For this task, the prompt is the benchmark prompt text itself, translated into the target language when needed. A representative prompt is:\par

{\ttfamily
Compose an email to a former professor or mentor asking for career advice. Briefly explain your current situation and ask for guidance on how to move forward.\par
}
\end{appendixexamplebox}

\section{Translation and Language-ID Audits}
\label{sec:measurement_audits}

\paragraph{Translated-prompt quality}
We score the $1{,}010$ NLLB-200 3.3B translations ($202$ language codes by five task templates) with the reference-free \texttt{wmt22-cometkiwi-da} model~\citep{rei-etal-2022-cometkiwi}. Table~\ref{tab:appendix_qe_audit} reports the resource-class breakdown over the $141$ languages shared with the PolyWrite analysis. Script variants of the same language are averaged before class aggregation. Quality is lower in classes~$0$ and~$1$, but it is not uniformly poor.

\begin{table}[!htbp]
\centering
\small
\begin{tabular}{lrrrrrr}
\toprule
Resource class & 0 & 1 & 2 & 3 & 4 & 5 \\
\midrule
Languages & 14 & 68 & 15 & 21 & 17 & 6 \\
Prompt translations & 70 & 340 & 75 & 105 & 85 & 30 \\
Mean COMETKiwi & .488 & .591 & .703 & .807 & .821 & .724 \\
\bottomrule
\end{tabular}
\caption{Reference-free quality estimation for NLLB-translated task templates by resource class.}
\label{tab:appendix_qe_audit}
\end{table}

Within classes~$0$--$1$, prompt quality only modestly correlates with downstream success across the five main models. Pearson/Spearman correlations are $.206/.304$ for code, $.275/.227$ for short stories, $.337/.304$ for teacher conversations, $.258/.267$ for translation to English, and $.039/.026$ for translation from English. PolyWrite backtranslation is also strong in these classes (class~$0$ \benchchrfpp{} mean/median $91.67/93.55$ over $52$ prompts; class~$1$ $92.87/94.19$ over $245$). A manual inspection of the lowest-QE class~$0$--$1$ templates found obvious repeated-token artifacts only for the Fon short-story and Yue translation-to-English templates; the worst remaining cases were generally partially untranslated rather than garbled. Translation artifacts therefore add noise, but do not explain the full low-resource gap.

\paragraph{GlotLID sensitivity}
We run GlotLID on ten local monolingual reference samples per language, treating their known codes as gold. The rebuttal-time audit covered $201$ languages/$2{,}010$ samples for Figure~\ref{fig:section3_takeaways} and recorded $81.4\%$ strict exact-code matches with $9.1\%$ no-decision; for the $137$ available PolyWrite languages/$1{,}370$ samples it recorded $86.5\%$ and $8.9\%$, respectively. Crucially, the lowest resource classes were not worse: class~$0$ obtains $90.0\%$ exact match and $3.6\%$ no-decision over $140$ samples, while class~$1$ obtains $88.5\%$ and $7.2\%$ over $650$. These strict figures can undercount semantically acceptable labels when GlotLID and the benchmark differ in dialect or script granularity; an initial inspection found about $13$ such language-level cases. GlotLID remains an imperfect single-tool proxy, so target-language-retention results should be interpreted with this caveat.

\section{Additional Robustness Results}

We report the LangID-confirmed support frontier because it is the closest thing in our setup to a usable behavioral notion of support.
This section adds two robustness checks from the current five-model comparison.
First, none of the current frontier models refuse any of the $20$ support-list prompts, so the main results are not being driven by refusal filtering.
Second, the broad models remain broad and the narrow models remain narrow even before LangID confirmation, but LangID confirmation removes a substantial number of unsupported raw claims for \modelgpt{}, \modelopus{}, and \modelmaverick{}.
Table~\ref{tab:appendix_support_robustness} reports the corresponding refusal counts and raw-versus-confirmed support-frontier bounds for all five current models.

\begin{table}[!hb]
\centering
\small
\begin{tabular}{lccccc}
\toprule
Model & Refusals & Raw $I/U$ & Confirmed $I/U$ & Raw $U$ & Confirmed $U$ \\
\midrule
\modelgpt{} & 0/20 & 28/187 & 25/134 & 187 & 134 \\
\modelopus{} & 0/20 & 38/187 & 33/140 & 187 & 140 \\
\modelgemini{} & 0/20 & 9/18 & 9/17 & 18 & 17 \\
\modelmaverick{} & 0/20 & 1/183 & 1/134 & 183 & 134 \\
\modelqwen{} & 0/20 & 7/41 & 7/35 & 41 & 35 \\
\bottomrule
\end{tabular}
\vspace{-0.4em}
\caption{Refusal-free support-list elicitation and the gap between raw elicited frontiers and LangID-confirmed frontiers for the current five-model comparison.}
\label{tab:appendix_support_robustness}
\vspace{-1em}
\end{table}

\noindent This pattern supports the main interpretation rather than weakening it.
For \modelgpt{}, \modelopus{}, and \modelmaverick{}, raw elicited support lists are even broader than the confirmed frontier, so the confirmed-support analysis is conservative.
For \modelgemini{} and \modelqwen{}, by contrast, LangID confirmation changes little because their elicited support frontiers are already relatively narrow.
The broader contradiction pool from which we selected the examples below is available in our public repository.\footnote{\url{https://github.com/antonisa/llm-languages}}

\section{Additional Contradiction Cases from the Current Frontier}

We present only a compact subset of contradictions in the core narrative.
This section adds more current-model cases organized by contradiction type.
Each box states the support claim or direct self-report, then shows the actual downstream prompt and output that violate that claim.
When the original prompt uses script that is not robust under the current pdflatex submission workflow, we show a transcribed or transliterated rendering of that prompt rather than the original glyph sequence.
All examples in this appendix come from the current five-model evaluation suite.

\subsection{Languages omitted from the claimed support frontier but still handled downstream}

\begin{appendixexamplebox}{GPT-5.4 excludes Acehnese from its claimed frontier but still translates an Acehnese prompt}
\textbf{Claimed support boundary.} Acehnese does not survive into GPT-5.4's LangID-confirmed support frontier.

\textbf{Downstream prompt.}
\begin{quote}\raggedright\footnotesize
Terjemahkan teks berikut ke bahasa Inggris. Simpan makna dan nada. Jawab dengan ``tidak dapat melakukan tugas ini'' jika itu kasusnya. Teks: Uroe nyoe uroe phon Nin di sikula.
\end{quote}

\textbf{Observed output.} \texttt{Today is Nin's first day at school.}

\textbf{Why this is contradictory.} The support frontier excludes Acehnese, yet the model still processes an Acehnese translation prompt successfully.
\end{appendixexamplebox}

\begin{appendixexamplebox}{GPT-5.4 excludes Balinese from its claimed frontier but still translates a Balinese prompt}
\textbf{Claimed support boundary.} Balinese is absent from GPT-5.4's LangID-confirmed support frontier.

\textbf{Downstream prompt.}
\begin{quote}\raggedright\footnotesize
Salin teks ring sor puniki ka basa Inggris. Ngiring simpang artos miwah nada. Saurin antuk ``unable to perform this task'' yéning punika sané dados jalaran. Teks: Dinane mangkin rahina kapertama Nin masekolah.
\end{quote}

\textbf{Observed output.} \texttt{Today is the first day of Nin going to school.}

\textbf{Why this is contradictory.} The claimed frontier leaves Balinese out, but the downstream translation behavior shows that the model still handles a Balinese input.
\end{appendixexamplebox}

\begin{appendixexamplebox}{Claude Opus 4.6 excludes Akan from its claimed frontier but still translates an Akan prompt}
\textbf{Claimed support boundary.} Akan is absent from Claude Opus 4.6's LangID-confirmed support frontier.

\textbf{Downstream prompt.}
\begin{quote}\raggedright\footnotesize
Sesa nsem a edidi so yi ko Borofo mu. Ma nteasee ne kasa no ntena ho. Se saa na ete a, fa ``ntumi nye adwuma yi'' bua. Nkyerewee: Enne ne sukuu da a edi kan a Nin wo.
\end{quote}

\textbf{Observed output.} \texttt{Today is Nin's first day of school.}

\textbf{Why this is contradictory.} The model's support frontier excludes Akan, but the multilingual evaluation still elicits a competent Akan-to-English translation.
\end{appendixexamplebox}

\begin{appendixexamplebox}{Claude Opus 4.6 excludes Albanian from its claimed frontier but still translates Albanian}
\textbf{Claimed support boundary.} Albanian does not appear in Claude Opus 4.6's LangID-confirmed support frontier.

\textbf{Downstream prompt.}
\begin{quote}\raggedright\footnotesize
Përkthejeni tekstin e mëposhtëm në anglisht. Ruajeni kuptimin dhe tonin. Përgjigjuni me ``nuk mund ta kryej këtë detyrë'' nëse është kështu. Teksti: Të dua.
\end{quote}

\textbf{Observed output.} \texttt{"I love you."}

\textbf{Why this is contradictory.} Albanian is excluded from the claimed frontier, yet the model still gives the correct translation for an Albanian sentence.
\end{appendixexamplebox}

\begin{appendixexamplebox}{Gemini 3.1 Flash-Lite excludes Albanian from its claimed frontier but still identifies it correctly}
\textbf{Claimed support boundary.} Albanian is outside Gemini 3.1 Flash-Lite's LangID-confirmed support frontier.

\textbf{Downstream prompt.}
\begin{quote}\raggedright\footnotesize
You are given a text snippet. Return the ISO 639-1 language code only. Text: Një mi kishte hyr në shtëpi.
\end{quote}

\textbf{Observed output.} \texttt{sq}

\textbf{Why this is contradictory.} The confirmed support frontier excludes Albanian, but the model still identifies an Albanian snippet with the correct code.
\end{appendixexamplebox}

\begin{appendixexamplebox}{Gemini 3.1 Flash-Lite excludes Asturian from its claimed frontier but still identifies it correctly}
\textbf{Claimed support boundary.} Asturian is outside Gemini 3.1 Flash-Lite's LangID-confirmed support frontier.

\textbf{Downstream prompt.}
\begin{quote}\raggedright\footnotesize
You are given a text snippet. Return the ISO 639-1 language code only. Text: Na nueche, depués de dormir, lloro.
\end{quote}

\textbf{Observed output.} \texttt{ast}

\textbf{Why this is contradictory.} Asturian is missing from the claimed support frontier, yet the model still recognizes an Asturian sentence correctly.
\end{appendixexamplebox}

\begin{appendixexamplebox}{Meta Llama 4 Maverick excludes Akan from its claimed frontier but still translates an Akan prompt}
\textbf{Claimed support boundary.} Akan does not appear in Meta Llama 4 Maverick's LangID-confirmed support frontier.

\textbf{Downstream prompt.}
\begin{quote}\raggedright\footnotesize
Sesa nsem a edidi so yi ko Borofo mu. Ma nteasee ne kasa no ntena ho. Se saa na ete a, fa ``ntumi nye adwuma yi'' bua. Nkyerewee: Enne ne sukuu da a edi kan a Nin wo.
\end{quote}

\textbf{Observed output.} \texttt{Today is Nin's first day at school.}

\textbf{Why this is contradictory.} The support frontier excludes Akan, but the downstream translation prompt still succeeds.
\end{appendixexamplebox}

\begin{appendixexamplebox}{Qwen 3.5 397B excludes Albanian from its claimed frontier but still translates Albanian}
\textbf{Claimed support boundary.} Albanian is absent from Qwen 3.5 397B's LangID-confirmed support frontier.

\textbf{Downstream prompt.}
\begin{quote}\raggedright\footnotesize
Përkthejeni tekstin e mëposhtëm në anglisht. Ruajeni kuptimin dhe tonin. Përgjigjuni me ``nuk mund ta kryej këtë detyrë'' nëse është kështu. Teksti: Të dua.
\end{quote}

\textbf{Observed output.} \texttt{I love you.}

\textbf{Why this is contradictory.} Albanian sits outside the claimed frontier, but the model still translates an Albanian sentence correctly.
\end{appendixexamplebox}

\subsection{Languages denied in direct self-report but still handled downstream}

\begin{appendixexamplebox}{GPT-5.4 says no to Tagalog but still translates Tagalog}
\textbf{Direct self-report.} Asked \texttt{Can you write or speak Q34057? Reply only with yes or no.}, GPT-5.4 answers \texttt{no}.

\textbf{Downstream prompt.}
\begin{quote}\raggedright\footnotesize
I-translate ang sumusunod na teksto sa Ingles. I-preserve ang kahulugan at tono. Tumugon sa ``hindi magawa ang gawaing ito'' kung iyon ang kaso. Text: Matutulog na ako.
\end{quote}

\textbf{Observed output.} \texttt{I'm going to sleep now.}

\textbf{Why this is contradictory.} The model explicitly denies support, then immediately translates a Tagalog prompt correctly.
\end{appendixexamplebox}

\begin{appendixexamplebox}{Claude Opus 4.6 says no to Acehnese but still handles an Acehnese translation prompt}
\textbf{Direct self-report.} Asked \texttt{Can you write or speak Acehnese? Reply only with yes or no.}, Claude Opus 4.6 answers \texttt{No}.

\textbf{Downstream prompt.}
\begin{quote}\raggedright\footnotesize
Terjemahkan teks berikut ke bahasa Inggris. Simpan makna dan nada. Jawab dengan ``tidak dapat melakukan tugas ini'' jika itu kasusnya. Teks: Uroe nyoe uroe phon Nin di sikula.
\end{quote}

\textbf{Observed output.} \texttt{The text is in Acehnese ... Here is the translation: Today is Nin's first day of school.}

\textbf{Why this is contradictory.} The direct self-report denies Acehnese, but the downstream response still recognizes the language and produces the translation.
\end{appendixexamplebox}

\begin{appendixexamplebox}{Claude Opus 4.6 says no to Bugis but still translates Bugis}
\textbf{Direct self-report.} Asked \texttt{Can you write or speak Bugis? Reply only with yes or no.}, Claude Opus 4.6 answers \texttt{No}.

\textbf{Downstream prompt.}
\begin{quote}\raggedright\footnotesize
Translate the following text to English. Preserve meaning and tone. Respond with ``unable to perform this task'' if that's the case. Text: Iyyaro esso pammulangnge ri sikolai Nin.
\end{quote}

\textbf{Observed output.} \texttt{That is the beginning of schooling for Nin.}

\textbf{Why this is contradictory.} The model denies Bugis in direct self-report, but then provides a Bugis-to-English translation.
\end{appendixexamplebox}

\begin{appendixexamplebox}{Meta Llama 4 Maverick says no to Welsh but still translates Welsh}
\textbf{Direct self-report.} Asked \texttt{Can you write or speak Welsh? Reply only with yes or no.}, Meta Llama 4 Maverick answers \texttt{No.}

\textbf{Downstream prompt.}
\begin{quote}\raggedright\footnotesize
Cyfieithu'r testun canlynol i Saesneg. Cadwch ystyr a thwn. Atebwch â ``unable to perform this task'' os dyna'r achos. Wyt ti'n byw yma?
\end{quote}

\textbf{Observed output.} \texttt{Do you live here?}

\textbf{Why this is contradictory.} The model directly denies Welsh support, but then translates a Welsh sentence correctly.
\end{appendixexamplebox}

\begin{appendixexamplebox}{Qwen 3.5 397B says no to Asturian but still translates Asturian}
\textbf{Direct self-report.} Asked \texttt{Can you write or speak Asturian? Reply only with yes or no.}, Qwen 3.5 397B answers \texttt{No}.

\textbf{Downstream prompt.}
\begin{quote}\raggedright\footnotesize
La traducción del testu siguiente al inglés. Preservar el significáu y el tonu. Responda con ``incapaz d'executar esta tarea'' si esti ye'l casu. El fueu siempre ye peligrosu.
\end{quote}

\textbf{Observed output.} \texttt{Fire is always dangerous.}

\textbf{Why this is contradictory.} The direct self-report denies Asturian support, but the model still translates an Asturian sentence correctly.
\end{appendixexamplebox}

\begin{appendixexamplebox}{Qwen 3.5 397B says no to Banjar but still translates Banjar}
\textbf{Direct self-report.} Asked \texttt{Can you write or speak Banjar? Reply only with yes or no.}, Qwen 3.5 397B answers \texttt{No}.

\textbf{Downstream prompt.}
\begin{quote}\raggedright\footnotesize
Tarjamahakan teks barikut ka bahasa Inggris. Simpan makna wan nada. Balas lawan ``kada kawa manggawi tugas ini'' amun kaya itu. Teks: Hari ini adalah hari pertama Nin di sekolah.
\end{quote}

\textbf{Observed output.} \texttt{Today is Nin's first day at school.}

\textbf{Why this is contradictory.} The model denies Banjar in self-report, then translates a Banjar prompt successfully anyway.
\end{appendixexamplebox}

\section{Additional Results}

\begin{figure}[!htbp]
    \centering
    \includegraphics[width=\textwidth]{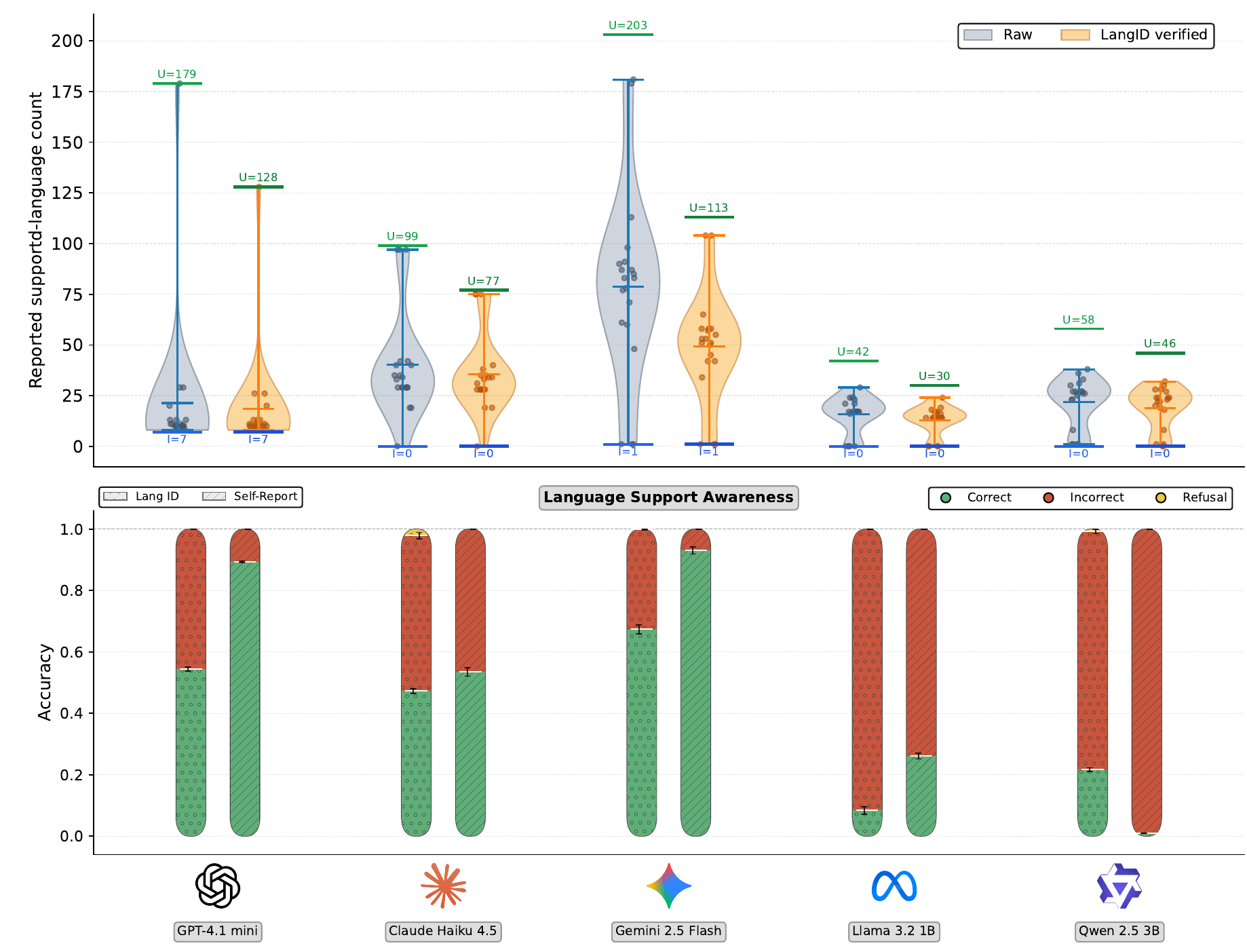}
    \vspace{-0.5em}
    \caption{Prompt-sensitive support claims and confirmed-support behavior for the earlier smaller-model comparison set: \modelgptmini{}, \modelhaiku{}, \modelgeminiflash{}, \modelllamaoneb{}, and \modelqwenthreeb{}. The same instability pattern from the main paper appears here too, but with even weaker confirmed intersections.}
    \label{fig:appendix_legacy_support}
    \vspace{-1em}
\end{figure}

\begin{figure}[!htbp]
    \centering
    \includegraphics[width=\textwidth]{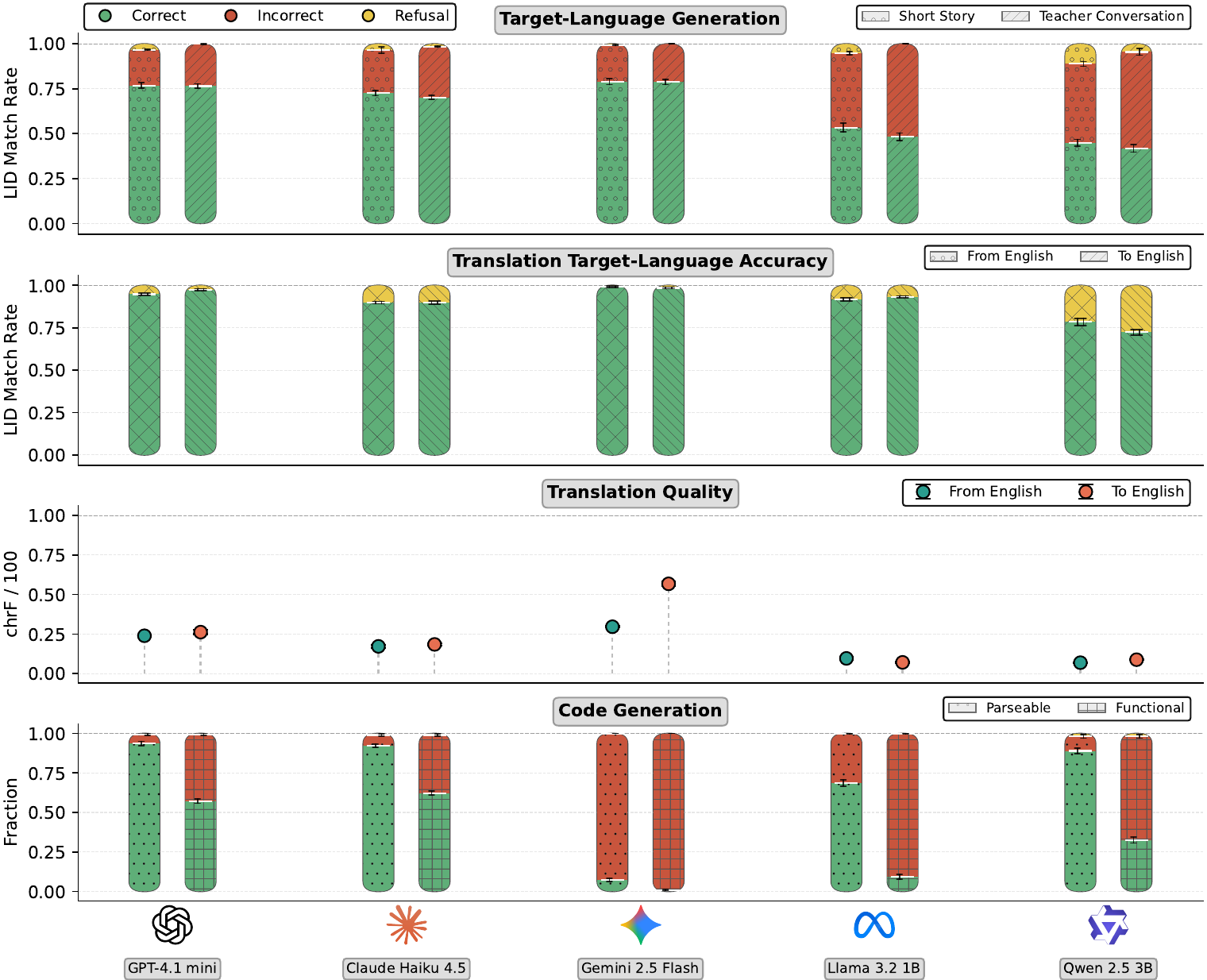}
    \vspace{-0.5em}
    \caption{Multilingual downstream benchmark for the earlier smaller-model comparison set. The same task-dependent ranking pattern remains visible, but absolute performance is generally lower than in the current frontier-model results shown in the main paper.}
    \label{fig:appendix_legacy_partb}
    \vspace{-1em}
\end{figure}

\begin{figure}[!htbp]
    \centering
    \includegraphics[width=\textwidth]{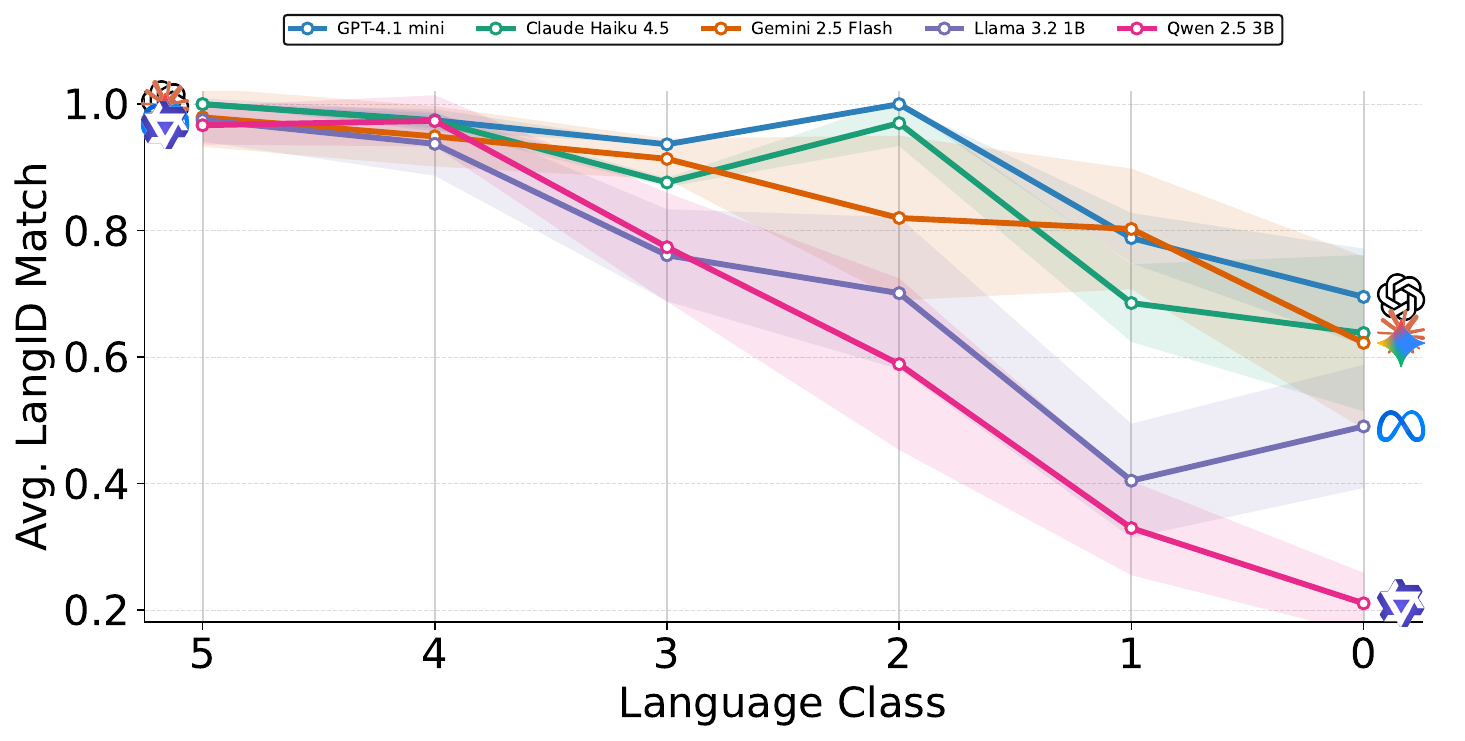}
    \vspace{-0.5em}
    \caption{Long-form target-language retention by \citep{joshi2020statefate} for the earlier smaller-model comparison set. The hardest low-resource classes remain the main separator, and the smallest open-weight models degrade especially sharply there.}
    \label{fig:appendix_legacy_taxonomy}
    \vspace{-1em}
\end{figure}

Figures~\ref{fig:appendix_legacy_support}, \ref{fig:appendix_legacy_partb}, and \ref{fig:appendix_legacy_taxonomy} provide the same support, downstream, and long-form-writing views for an additional smaller-model comparison set: \modelgptmini{}, \modelhaiku{}, \modelgeminiflash{}, \modelllamaoneb{}, and \modelqwenthreeb{}.
Table~\ref{tab:section3_support_probe} and Table~\ref{tab:section3_multilingual_results} provide the corresponding full numeric results for the support probe and downstream multilingual evaluation in the main comparison.
The qualitative picture matches the main-paper comparison, but the absolute numbers are usually worse.
Support-frontier instability is even more severe: in the LangID-confirmed support probe, \modelgptmini{} reaches only $I/U=7/128$, \modelgeminiflash{} reaches $1/113$, and \modelhaiku{}, \modelllamaoneb{}, and \modelqwenthreeb{} all collapse to zero confirmed intersection.
For the latter three models, the raw elicited intersection was already zero before GlotLID filtering. At least one prompt variant produced no extractable list, a refusal, or a sharply different support set; Haiku and Llama failures often involved problematic JSON-formatted outputs, while Qwen was especially prompt-sensitive. GlotLID reduces the union but does not cause the zero-intersection result.
The multilingual downstream benchmark also remains highly uneven.
\modelgptmini{} and \modelhaiku{} are clearly below \modelgpt{} and \modelopus{} on both translation and code, while the smallest open-weight models degrade sharply on the long-tail writing benchmark (class~$0$ drops to $0.49$ for \modelllamaoneb{} and $0.21$ for \modelqwenthreeb{}).
\modelgeminiflash{} is the main exception: it remains strong on X$\rightarrow$EN translation ($56.79$ \benchchrfpp{} in this comparison), yet it still exhibits unstable support declarations and very weak code functional correctness.
So the newer large-model comparison changes the absolute frontier, but it does not change the central conclusion: multilingual support remains unstable, over-claimed, and strongly task-dependent.

\begin{table}[t]
    \centering
    \scriptsize
    \begin{tabular}{lccccc}
    \toprule
    \textbf{Model} & \textbf{$I$} & \textbf{$U$} & \textbf{$I/U$} & \textbf{LangID} & \textbf{Self-report} \\
    \midrule
    \modelgpt{} & 25 & 134 & 0.187 & 0.583 & 0.995 \\
    \modelopus{} & 33 & 140 & 0.236 & 0.649 & 0.889 \\
    \modelgemini{} & 9 & 17 & 0.529 & 0.739 & 0.956 \\
    \modelmaverick{} & 1 & 134 & 0.007 & 0.590 & 0.640 \\
    \modelqwen{} & 7 & 35 & 0.200 & 0.557 & 0.700 \\
    \bottomrule
    \end{tabular}
    \vspace{-0.4em}
    \caption{Stability of self-declared support lists, alongside behavior on each model's confirmed support frontier. Higher self-report than LangID indicates that a model is more willing to claim support than to reliably stay in the requested language when asked to produce text.}
    \label{tab:section3_support_probe}
    \vspace{-1em}
\end{table}

\begin{table*}[t]
    \centering
    \scriptsize
    \resizebox{\textwidth}{!}{%
    \begin{tabular}{lcccccc}
    \toprule
    Model & \benchchrfpp{} EN$\rightarrow$X & \benchchrfpp{} X$\rightarrow$EN & Code parseable & Code functional & \benchpolywrite{} class 0 & \benchpolywrite{} class 1 \\
    \midrule
    \modelgpt{} & 35.44 & 45.46 & 0.994 & 0.760 & 0.739 & 0.883 \\
    \modelopus{} & 29.95 & 31.11 & 0.995 & 0.855 & 0.868 & 0.908 \\
    \modelgemini{} & 30.96 & 29.29 & 0.975 & 0.645 & 0.893 & 0.908 \\
    \modelmaverick{} & 27.58 & 24.34 & 0.987 & 0.651 & 0.696 & 0.864 \\
    \modelqwen{} & 24.99 & 31.06 & 0.869 & 0.594 & 0.765 & 0.713 \\
    \bottomrule
    \end{tabular}%
    }
    \vspace{-0.4em}
    \caption{Multilingual results across short-form generation, translation, code, and long-form writing. Translation columns report corpus-level \benchchrfpp{}, code columns use the canonical \benchmhumaneval{} alignment over the shared 203-item code family, and the final two columns report mean target-language retention in the two hardest resource buckets of the long-form writing benchmark.}
    \label{tab:section3_multilingual_results}
    \vspace{-1em}
\end{table*}

\end{document}